\documentclass[letterpaper]{article} % DO NOT CHANGE THIS
\PassOptionsToPackage{dvipsnames}{xcolor}
\usepackage{aaai2026}   % NOTE: right now for submitting to arxiv
\usepackage{times}  % DO NOT CHANGE THIS
\usepackage{helvet}  % DO NOT CHANGE THIS
\usepackage{courier}  % DO NOT CHANGE THIS
\usepackage[hyphens]{url}  % DO NOT CHANGE THIS
\usepackage{graphicx} % DO NOT CHANGE THIS
\usepackage{natbib}  % DO NOT CHANGE THIS AND DO NOT ADD ANY OPTIONS TO IT
\usepackage{caption} % DO NOT CHANGE THIS AND DO NOT ADD ANY OPTIONS TO IT
\usepackage{algorithm}
\usepackage{algorithmic}

\usepackage{newfloat}
\usepackage{listings}
\DeclareCaptionStyle{ruled}{labelfont=normalfont,labelsep=colon,strut=off} % DO NOT CHANGE THIS
\floatstyle{ruled}
\newfloat{listing}{tb}{lst}{}
\floatname{listing}{Listing}

\usepackage[utf8]{inputenc} % allow utf-8 input
\usepackage[T1]{fontenc}    % use 8-bit T1 fonts
\usepackage{url}            % simple URL typesetting
\usepackage{booktabs}       % professional-quality tables
\usepackage{amsfonts}       % blackboard math symbols
\usepackage{nicefrac}       % compact symbols for 1/2, etc.
\usepackage{microtype}      % microtypography
\usepackage[dvipsnames]{xcolor} % for using Goldenrod
\usepackage{soul}
\usepackage{longtable}
\usepackage{listings}
\lstdefinestyle{prompt}{
    basicstyle=\ttfamily\small,
    frame=none,
    breaklines=true,
    breakatwhitespace=true,
    breakindent=0pt,
    escapeinside={(*@}{@*)},
    numbers=none,
    numberstyle=\tiny,
    numbersep=3pt,
    xleftmargin=6pt,
    xrightmargin=6pt,
}
\usepackage{url}

\usepackage{amsmath,amsfonts,bm}

\usepackage{xspace}
\usepackage[flushleft]{threeparttable}
\usepackage{tablefootnote}
\usepackage{multirow}
\usepackage{hhline}
\usepackage{graphicx}
\usepackage{subcaption}
\usepackage{hanging}
\usepackage{color}
\usepackage{tikz}
\usetikzlibrary{calc,arrows,decorations.markings}
\usepackage{makecell, booktabs}
\usepackage{hhline}
\usepackage{ifthen}
\usepackage{comment}
\usepackage{cancel}

\newcommand{\RwR}{{\em SG-RwR}\xspace}
\def\task{{I}}
\def\gs{{\mathcal{S}}}
\def\sol{{\mathcal{A}}}
\def\gV{V}
\def\gE{E}
\def\query{q}
\def\code{h}
\def\anly{a}

\newcommand{\ReaLan}{Task Planner\xspace}
\newcommand{\ReaCoder}{Tool Caller\xspace}
\newcommand{\RetLan}{Verifier\xspace}
\newcommand{\RetCoder}{Code Writer\xspace}

\newcommand{\red}[1]{\textcolor{red}{{#1}}}
\newcommand{\green}[1]{\textcolor{green}{{#1}}}
\newcommand{\orange}[1]{\textcolor{orange}{{#1}}}
\newcommand{\yellow}[1]{\textcolor{Goldenrod}{{#1}}}
\newcommand{\gray}[1]{\textcolor{gray}{{#1}}}

\definecolor{agentColor}{HTML}{7FBC00}

\def\eqref#1{equation~\ref{#1}}
\def\1{\bm{1}}

\DeclareMathAlphabet{\mathsfit}{\encodingdefault}{\sfdefault}{m}{sl}
\SetMathAlphabet{\mathsfit}{bold}{\encodingdefault}{\sfdefault}{bx}{n}

\def\gG{{\mathcal{G}}}

\usepackage{supertabular, booktabs}
\usepackage{float}
\usepackage{placeins}
\renewcommand{\RwR}{{\em SG\textsuperscript{2}}\xspace}
\nocopyright        % NOTE: for submitting to arxiv
\title{Schema-Guided Scene-Graph Reasoning based on Multi-Agent Large Language Model System}
\author{
    Yiye Chen\textsuperscript{\rm 1},
    Harpreet Sawhney\textsuperscript{\rm 2},
    Nicholas Gyd\'{e}\textsuperscript{\rm 2},
    Yanan Jian\textsuperscript{\rm 2},
    Jack Saunders\textsuperscript{\rm 3}, \\
    Patricio Vela\textsuperscript{\rm 1},
    Ben Lundell\textsuperscript{\rm 2}
}
\affiliations{
    \textsuperscript{\rm 1}Georgia Tech \\
    \textsuperscript{\rm 2}Microsoft \\
    \textsuperscript{\rm 3}University of Bath

    \{yychen2019, pvela\}@gatech.edu,  jrs68@bath.ac.uk \\
    \{harpreet.sawhney, gydenicholas, yananjian, benjamin.lundell\}@microsoft.com,

}

\usepackage{bibentry}
\begin{document}

\maketitle

\begin{abstract}
% Situated LLM reasoning is important, and use scene graph is one way.
    % Grounding the reasoning and planning capabilities of Large Language Models (LLMs) in specific environments remains a significant challenge. 
    % Recent advancements have demonstrated the effectiveness of representing environments as scene graphs, which offer a flexible and structured way to encode diverse semantic and spatial information.
    Scene graphs have emerged as a structured and serializable environment representation for grounded spatial reasoning with Large Language Models (LLMs).
% Existing method has drawback
    % However, existing approaches that naïvely prompt LLMs with serialized graph representations often suffer from hallucinations when processing large graphs and fail to produce graph-grounded reasoning steps for complex spatial problems.
% What we propose
    % In this work, we propose \textbf{\RwR}, an iterative scene graph reasoning framework that addresses these limitations through \textit{scene graph schema} prompting.
    % In this work, we propose \textbf{\RwR}, an iterative scene graph reasoning framework based on scene graph \textit{schema} prompting and the code-writing capacity of LLMs.
    In this work, we propose \textbf{\RwR}, an iterative \textbf{S}chema-\textbf{G}uided \textbf{S}cene-\textbf{G}raph reasoning framework based on multi-agent LLMs.
    The agents are grouped into two modules: a (1) \textit{Reasoner} module for abstract task planning and graph information queries generation, and a (2) \textit{Retriever} module for extracting corresponding graph information based on code-writing following the queries.
    Two modules collaborate iteratively, enabling sequential reasoning and adaptive attention to graph information.
    The scene graph schema, prompted to both modules, serves to not only streamline both reasoning and retrieval process, but also guide the cooperation between two modules.
    This eliminates the need to prompt LLMs with full graph data, reducing the chance of hallucination due to irrelevant information.
    % Unlike prior works, both agents are prompted only with the \textit{scene graph schema} rather than the full graph data, which reduces the hallucination by limiting input tokens, and drives the Reasoner to generate reasoning trace abstractly.
    % , ensuring close alignment between the reasoning and retrieval processes.
    % facilitates focused attention on task-relevant graph information and enables sequential reasoning on the graph essential for complex tasks.
    % Following the trace, the Retriever programmatically query the scene graph data based on the schema understanding, allowing dynamic and global attention on the graph that enhances alignment between reasoning and retrieval. 
    % Additionally, the code-writing design allows tool-using to solve problems beyond the capacity of LLMs, which further enhance its reasoning ability facing complex tasks.
% Summarize results
    Through experiments in multiple simulation environments, we show that our framework surpasses existing LLM-based approaches and baseline single-agent, tool-based Reason-while-Retrieve strategy in numerical Q\&A and planning tasks.
    % , and can benefit from task-level few-shot examples, even in the absence of agent-level demonstrations.
% code
    % Project code will be released. %upon acceptance.
\end{abstract}

% Uncomment the following to link to your code, datasets, an extended version or similar.
% You must keep this block between (not within) the abstract and the main body of the paper.
% \begin{links}
%     \link{Code}{https://aaai.org/example/code}
%     \link{Datasets}{https://aaai.org/example/datasets}
%     \link{Extended version}{https://aaai.org/example/extended-version}
% \end{links}

\section{Introduction}
\label{sec:intro}

%%%%% figure %%%%%%%%%%%%%%%%%%%%%%%%%%%%%%%%%%%%%%%%%%%%%%%%%
\begin{figure*}[t!]
 
  % \vspace*{-0.1in}
  \centering
  \scalebox{0.95}{
    \begin{tikzpicture}
     \node[anchor=north west] at (0in,0in)
      {{\includegraphics[width=1.0\textwidth,clip=true,trim=0
      280 0 0]{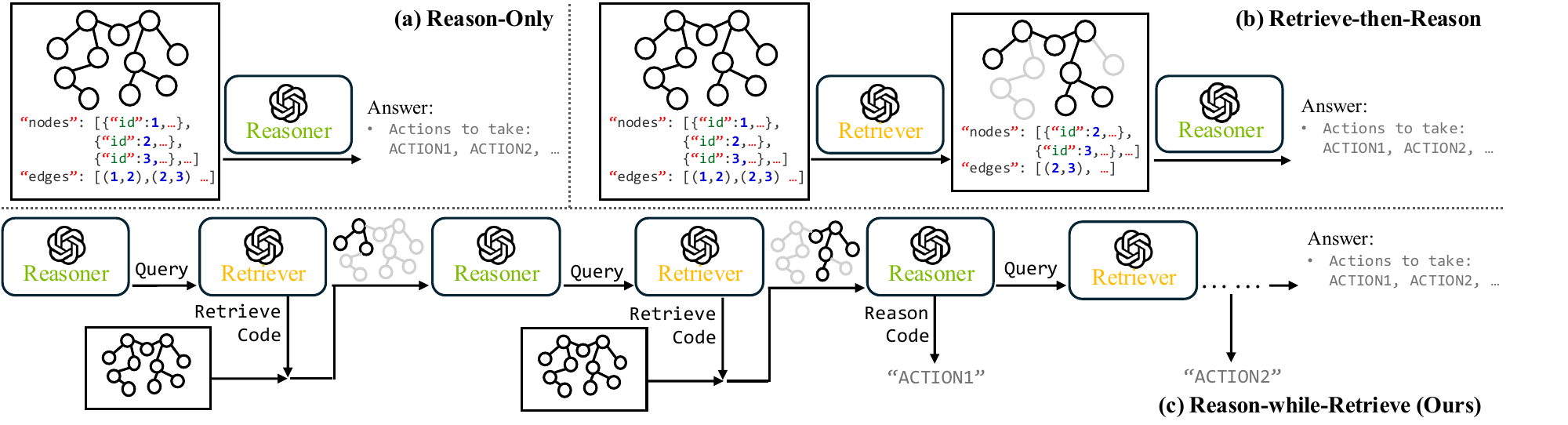}}};
%     \node[yshift=-0pt,anchor=north west] at (0.1in,0.0in) {\bf \small (a)};
%     \node[anchor=north west] at (0.92in,-0.05in) {\textbf{(a)}};
%     \node[anchor=north west] at (2.00in,-0.05in) {\textbf{(b)}};
%     \node[anchor=north west] at (3.09in,-0.05in) {\textbf{(c)}};
    \end{tikzpicture}
  }
 \vspace*{-0.1in}
  \caption{\textbf{LLM Graph Processing Framework Comparison}.
  (a) Reason-Only: A Reasoner is directly prompted with a full textualized graph.
  (b) Retrieve-then-Reason: A Retriever filters out a task-related sub-graph followed by another Reasoner processing the remaining of the graph.
  (c) Reason-while-Retrieve (Ours): A Reasoner and a Retriever collaborate in solving a task by attending to the graph dynamically based on the progress in solving the task.
  }
  \vspace*{-0.1in}
 \label{fig:Overview}
\end{figure*}
%%%%%%%%%%%%%%%%%%%%%%%%%%%%%%%%%%%%%%%%%%%%%%%%%%%%%%%%%%%%%

%%% Why reasoning on scene graphs are meaningful
% \Ben{We need to update the references here.}

%With the remarkable prowess in language interpretation and reasoning \citep{gpt4, llama2}, Large Language Models (LLMs) have been increasingly adopted in the embodied planning tasks \citep{huang2023embodied, innermono, socratic}, including plan generation \citep{LLMPlanner}, interaction \citep{joublin2024copal}, and action selection \citep{sayplan}.
%Despite the progress, the challenge of grounding the LLMs reasoning to situated environments remains unsolved, primarily due to the absence of a environmental representation that LLMs can effectively process \citep{groundedDecoding}. 
%While LLMs can interface with external tools to directlyx process perceptual data such as images \citep{CaP, voxposer}, they are unable to comprehend the intermediate, non-textual outputs from those tools, which prohibits generating the grounded reasoning trace.
%In contrast, \textbf{scene graphs} represent environments as hierarchical graphs that encapsulate spatial relationships and semantic attributes in a structured and serializable format \cite{hierarchicalSg, hydra}. 
%As a result, scene graphs have emerged as a scalable, high-level environment representation for LLM-based spatial reasoning and planning, showing effectiveness in both simulation-based \cite{octopus} and real-world applications \cite{sayplan, conceptgraphs, GRID, spatialrgpt}. 

With Large Language Models (LLMs) showing remarkable prowess and versatile skills across a wide range of domains, recent research has increasingly focused on grounding the LLMs their reasoning in situated environments.
Scene graphs have emerged as a scalable, high-level environment representation for LLM-based spatial reasoning and planning, showing effectiveness in both simulation-based \cite{octopus} and real-world applications \cite{conceptgraphs, GRID, spatialrgpt}.
Prior work has explored graphs-as-text as the LLM input for the single generation with various prompt guidance \cite{talkLikeGraph, conceptgraphs}, categorized as \textit{"Reason-only"} methods in Figure. \ref{fig:Overview}.
A more advanced strategy, \textit{"Retrieve-then-Reason"} \citep{reasonOnGraph, thinkOnGraph, sayplan}, 
improves upon this by using the LLM agent to trim the graph first for retaining only task-relevant subgraph before reasoning.
Despite the effort, LLMs still frequently hallucinate wrong solutions~\citep{NLGraph, sayplan},
underscoring the need for continued research on the intersection of LLMs and scene graphs.
 % Yet, they are prone to hallucinations or exceed token limits when handling large graphs \citep{NLGraph}. 
% Despite the progress, the challenge of grounding the LLM's reasoning to a scene graph remains unsolved, primarily due to the absence of a representation of the data that LLMs can effectively process \citep{groundedDecoding}.\Ben{<- This might not be the right reference anymore.}
We argue that this difficulty stems from a well-documented property of LLMs: that they are easily distracted by redundant information~\citep{hallRedundantContext1, hallRedundantContext2}.
This limitation is problematic in spatial tasks, where the reasoning process involves sequential, step-wise attention shifts over the graph.
It suggests that the majority of graph data could be irrelevant at any intermediate step of reasoning,
which might degrade LLMs' performance.
% Specifically, the highly-structured and hierarchical nature of a scene graph sequential and dynamically shifting attention required by spatial reasoning tasks, means that \textit{much of the scene graph data is irrelevant at any intermediate step of reasoning}.
% Thus, we explore 
% \red{\textbf{Originally}:
% In this work, we argue that this difficulty stems from a well-documented property of LLMs: They are easily distracted by irrelevant information CITE.
% Specifically, the highly-structured, hierarchical nature of a scene graph coupled with the local (or semi-local) nature of most spatial tasks, means that, for any given task, \textit{much of the data in the scene graph is irrelevant}.
% }
% Irrelevant information has been demonstrated to be particularly deleterious to LLMs during reasoning tasks CITE, and so
% Thus, 
% we present an approach that enables structured access to the scene graph and thereby filters out irrelevant information for the reasoning trace.

A promising solution to the problem is the \textit{Reason-while-Retrieve} strategy, which enables dynamic attention on the information by iteratively carrying out the two steps~\citep{activeRAG, press2022measuring}. 
We first explore an iterative solution based on ReAct~\citep{react}, where scene graph API(s) are curated as graph data access "actions" to enable interleaved task solving and graph information retrieval. 
While this solution is effective, we observe that it is highly sensitive to the design of available APIs (or tools). 
In particular, fixed-capacity APIs severely constrain the access patterns on the graph.
When the API set lacks expressiveness, the agent is forced to invoke more API calls, resulting in inefficient exploration on the graph.
What's more, this inefficiency is amplified by the single-agent nature of ReAct, where the retrieval and reasoning is not decoupled. Since the entire task solving history is re-prompted back to the agent in a loop,
redundant context accumulation can impair future reasoning or retrieval steps \cite{chiang2024overreasoning, wu2024how}.
To mitigate this issue, we develop a \textbf{S}chema-\textbf{G}uided \textbf{S}cene-\textbf{G}raph reasoning approach based on multi-agent architecture, (dubbed \textbf{\RwR}). 
The framework is comprised of two modules:
a \textit{Reasoner} module that decomposes the task and generates information queries for subsequent steps; and a \textit{Retriever} module that processes the queries and retrieves related graph information for the Reasoner.
One key innovation in our approach is the incorporation of \textit{scene graph schema}.
Treating a scene graph as a specific instance of an abstract schema, we frame the role of the Retriever as \textit{graph database query language generation}—that is, translating the natural language queries into executable graph programs.
The query language is then executed on the scene graph to obtain the queried information thereby filtering out irrelevant data.
Furthermore, the schema is also prompted to the Reasoner for  support abstract, structure-aware reasoning, and to guide the generation of schema-aligned natural language queries for the Retriever.
The multi-agent design is essential for the autonomous operation of our system: where the inter-module split ensures explicitly separation of the reasoning and retrieval, while intra-module agents collaborate to enhance the reliability of both reasoning and programming-based retrieval.

We evaluate our method with two simulation environments: BabyAI \citep{babyai}, a 2D grid world environment; and VirtualHome \citep{virtualhome}, a large-scale indoor multi-room environment.
Our experiments on numerical Q\&A and planning tasks show that \RwR greatly improves the reasoning ability of LLMs on scene graphs, 
outperforming ReAct and graph prompting baselines in all evaluations. 
What's more, we show that even being constrained to the same static APIs with limited functionalities, our multi-agent framework still achieves better results compared to ReAct.
In summary, our contributions include:
\begin{itemize}
   \item Exploring Reason-while-Retrieve framework with reasoning-oriented information gathering mechanism for task solving on scene graphs. 
   \item A schema-based multi-agent approach that enables dynamic information retrieval with LLM programming and decoupled reasoning and retrieval.
   \item Showing the efficacy of the proposed \RwR, which achieves superior performance in two distinct environments that encompass a wide range of tasks.
\end{itemize}

\section{Related Literature}
\label{sec:lit}

\paragraph{Language models for Task and Motion Planning}
% With the advance of large language or multimodal models, many earlier
Many existing efforts harness the power of large language models for decision making \cite{xi2023rise, leap, llm+p} and robotic control \cite{plan-seq-learn, zhang2023bootstrap, text2motion, CGNet, hatori2018interactively}. With rich built-in knowledge and in-context learning ability,
%trained from the large internet-scale text corpora, 
language models are used for generating task-level plans \cite{raman2022planning, gao2024physically}, action selection \cite{saycan, pivot}, processing environmental or human feedback \cite{CLAIRIFY}, training or finetuning language-conditioned policy models \cite{octo, rt-x, szot2023large}, and more. 
To factor in the environment during planning, recent studies have explored using LLMs for programmatic plan generation \cite{progprompt}, combining knowledge from external perception tools \cite{CaP, voxposer} or grounded decoding \cite{groundedDecoding}, and value function generation \cite{language2reward}. While proven effective, those methods are limited to small scale environments, and rely on expert perception models to extract task-related states from the scene representation with implicit spatial structure. In this work, we study using pretrained LLMs to process the the global representation of large environments with explicit structure.%, and generate the solution that is grounded in the environment. 

\paragraph{Graph as the Scene Representation}
The scope of the solvable task is largely determined by the state representation. Compare to sensory representation such as images or point clouds, scene graphs are compact thus scalable to large environments \cite{greve2024collaborative}, structured to represent spatial layout explicitly \cite{hydra, wu2021scenegraphfusion}, and efficient in representing diverse states of the environment \cite{3dsg}. Therefore, they have been used in various manipulation or navigation tasks \cite{3dsgNav, hierarchicalSg}. In this paper, we exploit these favorable features of the scene graph representation to ground the reasoning process of LLMs to the environment. 

\paragraph{LLMs for Reasoning on Graph}
Leveraging language models to reason with graphs is a growing area. While prior works trains to integrates graph and language knowledge \cite{instructGLM, GRID}, recent study explores serializing graph-structured data as prompts for pretrained LLMs \cite{NLGraph, talkLikeGraph}. This strategy has been successfully used in knowledge-graph-enhanced LLMs reasoning \cite{thinkOnGraph, reasonOnGraph} and scene-graph-based robotic task planning \cite{conceptgraphs}. Closest to our work, SayPlan \cite{sayplan} prompts scene graphs to LLMs and designs a Retrieve-then-Reason framework for robotic planning. However, its room-by-room retrieval heuristic is only effective in the object search task. Instead, we design the \RwR framework for general spatial reasoning with scene graphs.

%We further incorporate the code-writing and tool-use ability to LLMs, so that our proposed method can effectively retrieve information based on scene graphs and address numerical tasks that fall beyond the expertise of LLMs \cite{AliceInWonderland}.
\section{Method}

%%%%% figure %%%%%%%%%%%%%%%%%%%%%%%%%%%%%%%%%%%%%%%%%%%%%%%%%
\begin{figure*}[t!]
 
  % \vspace*{-0.1in}
  \centering
  \scalebox{0.97}{
    \begin{tikzpicture}
     \node[anchor=north west] at (0in,0in)
      {{\includegraphics[width=1.0\textwidth,clip=true,trim=0
      100pt 0 0]{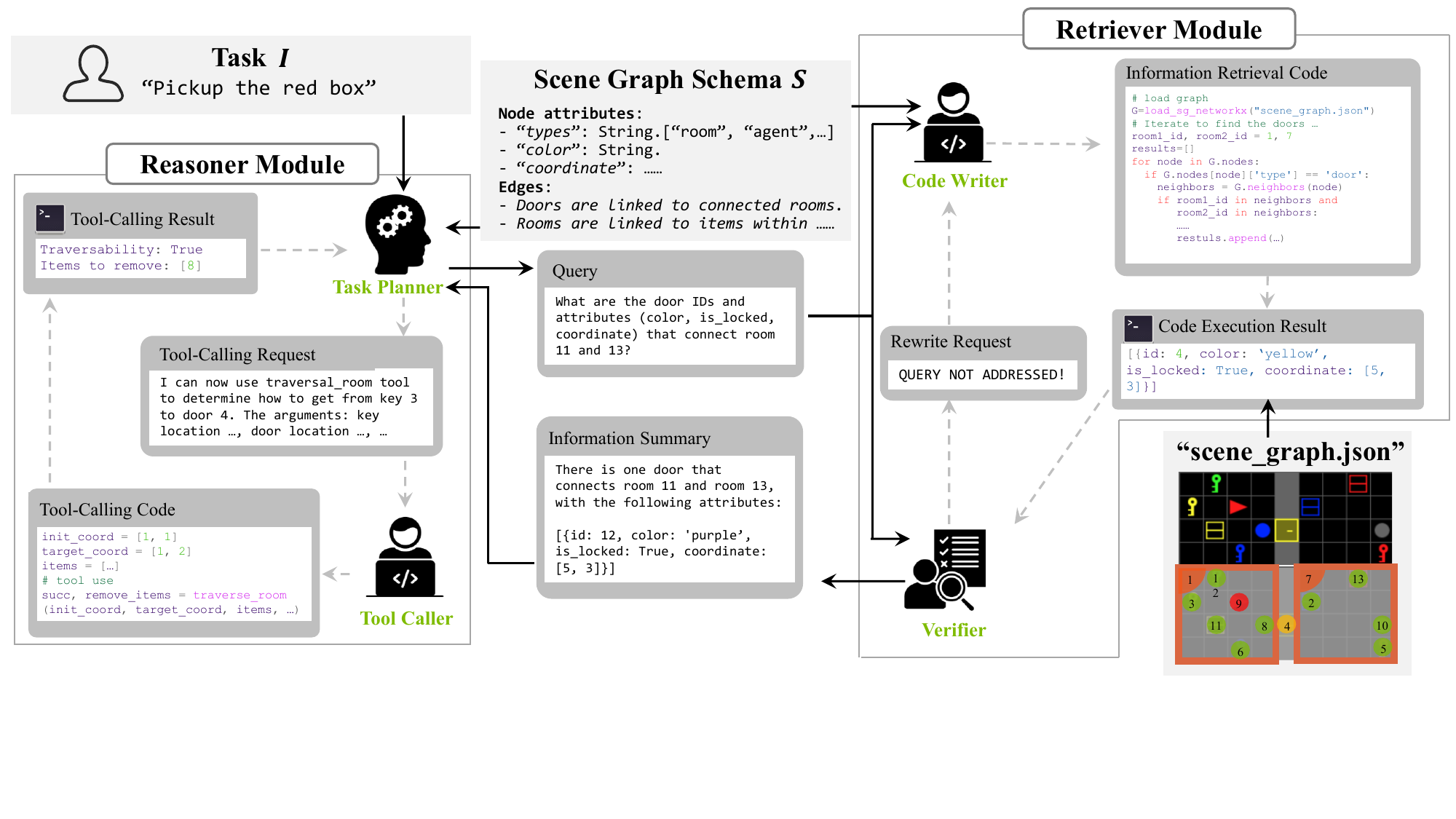}}};
%     \node[yshift=-0pt,anchor=north west] at (0.1in,0.0in) {\bf \small (a)};
%     \node[anchor=north west] at (0.92in,-0.05in) {\textbf{(a)}};
%   \node[anchor=north west] at (2.00in,-0.05in) {\textbf{(b)}};
%     \node[anchor=north west] at (3.09in,-0.05in) {\textbf{(c)}};
    \end{tikzpicture}
  }
  % \vspace*{-0.05in}
  \caption{\textbf{\RwR Workflow}.
  The method solves tasks on scene graphs through the iterative collaboration of two LLM-based multi-agent modules: the Reasoner and the Retriever. 
  The \textit{Scene Graph Schema} lies at the core of the approach, guiding the Reasoner to break down the problem and formula queries, and enabling the Retriever to generate code that efficiently processes the scene graph and fulfills those queries. 
  \textcolor{agentColor}{LLM agents} are color coded. 
  % It solves tasks on scene graphs through the cooperation of two LLM-based multi-agent modules: a Reasoner module that progressively raises queries and reasons based on accumulated information, and a Retriever module that writes code to process the scene graph to fulfill the queries. 
  % The scene graph schema is prompted to synergize the reasoning and retrieval.
  % Additionally, both agents employ the code-writing skill: Retriever programs to retrieve graph information based on the schema, while the Reasoner writes code to utilize external tools for solving complex atomic problems. In the graph, 
  % \protect{\raisebox{-.05cm}{\includegraphics[height=.30cm]{figs/code.png}}} and 
  % \protect{\raisebox{-.05cm}{\includegraphics[height=.30cm]{figs/code_exe.png}}}
  % represent code writing and execution, respectively.
  % They retrieve graph information $\bm{\gG}^\prime$ or enhance the analysis $\bm{\anly}$.
  }
  \vspace*{-0.15in}
 \label{fig:Method}
\end{figure*}
%%%%%%%%%%%%%%%%%%%%%%%%%%%%%%%%%%%%%%%%%%%%%%%%%%%%%%%%%%%%%

\subsection{Problem Statement}
Our problem setting involves a natural language task instruction $\task$ and a scene graph $\gG = (\gV, \gE)$, where $\gV$ and $\gE$ denote vertices and edges, respectively. Each node $\gV_i$ represents an item with its attributes, such as coordinates or colors, while each edge indicates a type of spatial relationship, such as inside or on top of. Additionally, we assume access to the \textit{scene graph schema} $\gs$, which is a textual description of types, formats, and the semantics of the graph vertices and edges. Our objective is to generate the solution of $\task$ using LLMs, based on the available information above, expressed as $\sol = f(\task, \gG, \gs; LLMs)$.

\subsection{Overview of \RwR}
We explore grounding the task solving to scene graphs \textbf{based on the scene graph schema $\gs$}.
We develop \RwR, a schema-guided multi-agent framework that iteratively reasons through the next steps and retrieves the necessary information from the graph.
As shown in Figure \ref{fig:Method}, our method contains two multi-agent modules: a \textit{Reasoner} and a \textit{Retriever}. 
Both modules consist of two LLM agents: the Reasoner is comprised of a \ReaLan and a \ReaCoder, whereas the Retriever is comprised of a \RetCoder and a \RetLan.
% The system initializes with the Scene Graph Schema, the Environment Description, general Guidance to direct the cooperation process, and task-dependent information such as the description of Agent Actions and Reasoning Tools. 
Given a task, the Reasoner determines the next substep to approach the task and identifies necessary scene graph information. It then raises a natural language query to the Retriever for this information. Upon receiving the query, the Retriever processes the scene graph through code-writing and sends the data back to Reasoner. By iteratively performing these steps, both modules collaborate to solve the task. 
Prompted with the schema $\gs$, both modules in our approach are NOT contextualized with the graph data $\gG$, which differs from prior graph reasoning methods.
Formally, at each time step $t$:
\begin{align}
    \anly_t, \query_t &= Reasoner(\{\anly_0, \query_0, \gG^\prime_0\}, \{\anly_{1}, \query_{1}, \gG^\prime_{1}\}, \cdots; \bm{\gs}) \\
    \code_t &= Retriever(\query_t; \bm{\gs}) \\
    \gG^\prime_{t} &= \code_t(\gG)
\end{align}
where $\anly$ denotes the Reasoner's internal analysis; $\query$ represents queries for the graph information; $\code$ denotes the retrieval program following the query; and $\gG^\prime$ refers to the retrieved information by executing the code on the scene graph $\gG$.

%\cite{react, iterRG}
Importantly, our method differs from previous iterative methods such as ReAct \cite{react} in the following ways: 
(1) Our method is conditioned on the schema input instead of API annotations;
(2) Our method programs to retrieve information instead of relying on provided APIs, which is more flexible and efficient;
(3) Our method, powered by multi-agent designs, separates the reasoning the graph exploration processes. 
As we show in the experiment section, these designs improve the efficacy of our approach and robustness against limited API capacity.

The remaining of the section describes the multi-agent workflow as well as the roles of each agent. For the detailed prompt of each agent, please see the supplementary material.

% , the two agents in \RwR only exchange the query $\query$ and the corresponding graph data $\gG^\prime$, excluding the underlying thought process, such as $\anly$ and $\code$. 
% As we will show, this agent-level context filtering, enabled by our two-agent design, is critical for eliminating the interference from irrelevant conversation history, thereby ensuring a seamless and automated cooperative task-solving process.

% including the {\em Explanation} of intermediate conclusions in language and the {\em Reasoning Code} for tool using or sub-task solving;

% Our system initializes with the Scene Graph Schema, the Environment Description, general Guidance to direct the cooperation process, and task-dependent information such as the description of Agent Actions and Reasoning Tools. Then, given the Task, the Reasoner outputs analysis in natural language labeled as {\em Explanation}, and {\em Query} the Retriever. In turn, given the Scene Graph and a Query, the Retriever provides structured responses grounded in the Scene Graph. This process iterates until the Reasoner outputs a plan.

% The next subsections explain workflows of each agent, as well as techniques that ensure a fluent and automated task-solving process.

\subsection{Reasoner Module}

% Reasoner is the central agent steering the task-solving iterations. We prompt it with the schema $\gs$, environment and task information (such as action description for the planning task), annotations of reasoning tools, general guidance to ensure automated task-solving conversation, and optionally, few-shot task-level examples. Reasoner then initiates the conversation with Retriever to solve a given task.
The \textbf{\ReaLan} is the central agent steering the task-solving iterations based on the scene graph schema $\gs$. It takes as input the task $\task$ and schema $\gs$ and initiates the problem solving process. 
Initially, without any knowledge about the graph data, \ReaLan analyzes $\task$ and $\gs$, 
% generates the first analysis $\anly_0$, 
and sends out the first associated query $\query_0$ to the Retriever. At the $t^{th}$ round of conversation, it consumes past analyses, queries, and retrieved information, and then generates one of the three types of responses, each of which is sent to a different recipient agent:
(1) QUERY: querying for more information from the Retriever. This response is sent to the Retriever-side \RetCoder;
(2) TOOL-CALL: calling a reasoning tool to process collected information. This response is sent to \ReaCoder;
(3) SOLUTION: generating the solution, which terminates the task solving process. 
% It then generates the next corresponding analysis $\anly_t$ and query $\query_{t}$, where $\anly_t$ involves intermediate conclusions and the next subtask to be solved, which informs and justifies $\query_{t}$.
% NOTE: can simplify below.
The TOOL-CALL invokes provided reasoning tools to solve complex spatial sub-problems. This is motivated by previous literature revealing the inability of to reliably solve quantitative problems \citep{llmMathReason}. To circumvent the deficiency, we follow prior work \citep{toolformer, ART} to enable tool-use by providing \ReaLan with annotations of programmatic functions, 
such as \texttt{\small traverse\_room} for solving navigation problem in Figure.~\ref{fig:Method},
so that it is able to suggest suitable tools and corresponding arguments to address atomic problems critical to the given task family. 

Concretely, we prompt \ReaLan to generate the following outputs at each step:

\begin{description}
    \item \textbf{Explanation}: Summarize the reasoning process and justify the generation of the current response.
    \item \textbf{Mode}: The type of the current response.
    \item \textbf{Content}: The detailed message in the current response, such as the desired graph information for QUERY; tool name and arguments for TOOL-CALL; or the final answer for SOLUTION.
\end{description}
We filter out only the \textbf{Content} message for the recipient agent, reducing the interference by the redundant reasoning process behind the request. Few-shot examples are prompted to \ReaLan to enhance the performance in task solving and enforce the output format. 

% emphasizing the importance of schema
The schema prompt $\gs$ is critical by serving two purposes. First, it leads the \ReaLan to reason the task \textit{abstractly}, reducing hallucination due to task-irrelevant graph information \citep{NLGraph}. What's more, it streamlines the Reasoner-Retriever collaboration by guiding the generation of query message, ensuring that it is parsable by the Retriever. 

The \textbf{\ReaCoder} is prompted with the reasoning tool annotations, and is responsible for translating the tool-calling messages from \ReaLan into executable python programs. 
We observe that \ReaLan along might not be able to invoke tools with the correct format.
Hence we split the burden and use a separate \ReaCoder agent for formatting. 
While other structured output techniques exist~\citep{structureOutput}, we find adding another agent to be most flexible and sufficient for our application.

\subsection{Retriever Module}

The \textbf{\RetCoder} processes the query $\query$ from Reasoner and, conditioned on the graph schema prompt $\gs$, generates the code $\code$ to process scene graphs programmatically.
The schema and the query language enable the \RetCoder to compose low-level APIs, introduce control-flow, and generally convert the natural language queries into executable program to run on the scene graph.
This code-writing strategy offers significant advantages over traditional API-calling methods.
By enabling efficient graph traversal for query-oriented information filtering, the irrelevant parts of the graph never enter the retriever's context window and so retrieved information is well-aligned with the reasoning demands.

Following prior work~\citep{selfdebug}, we incorporate a self-debugging mechanism to address possibility that LLMs may generate unexecutable code even with sufficient context. 
Specifically, we iteratively re-prompt the past code-writing attempts and execution errors back to \RetCoder until successful execution is achieved. 

Even with error prevention mechanism, the final code execution might not produce valid results for the query due to multiple possibilities, such as missing result output or scattered graph information output along the multiple code-rewriting attempts.
To mitigate the issue, we introduce the \textbf{\RetLan} agent to evaluate the code execution results.
It takes as input the information retrieval query as well as all past code execution results, and determines if the query is addressed. 
If the query is deemed addressed, it prompts the \RetCoder to re-write the code. Otherwise, it summarizes the result and sends it back to the Reasoner.

Note unlike the single-agent ReAct method, our multi-agent pipeline naturally filters the context exploiting the conditional independence structure inherent in the task solving process. For example, \RetCoder generates the program $\code$ solely based on the query $\query$ without full reasoning history, and the \ReaLan receives only the retrieved graph data $\gG^\prime$ without being exposed to code generation and correction details. 
This ensures that each agent operates strictly within their designated responsibility, free from irrelevant distractions that could impair their response \cite{yoran2023making}. 

% \Ben{I think this whole Method Section need to emphasize the various ways in which our architecture filters out irrelevant information by design. The retriever stops the reasoner from seeing unrelated parts of the graph. Given the query q and the schema S, the response G' is independent of the rest of the reasoning trace. I think we should make this Markovian property very explicit. Basically, if this is now the main justification for our approach, we really need to hammer it home here.}

%%%%% figure %%%%%%%%%%%%%%%%%%%%%%%%%%%%%%%%%%%%%%%%%%%%%%%%%
\begin{figure*}[t!]
 
  % \vspace*{-0.1in}
  \centering
  \scalebox{0.8}{
    \begin{tikzpicture}
     \node[anchor=north west] at (0in,0in)
      {{\includegraphics[width=1.0\linewidth,clip=true,trim=0
      220pt 60pt 0]{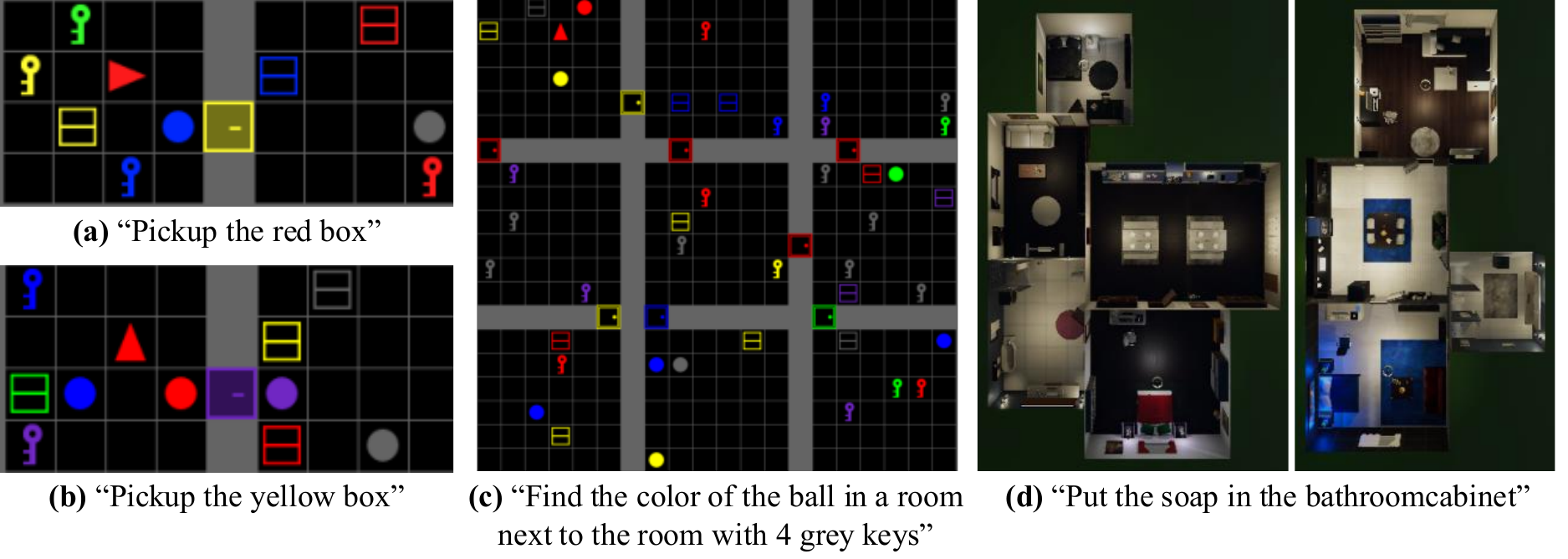}}};
    \end{tikzpicture}
  }
  \vspace*{-0.1in}
  \caption{\textbf{Experiment Settings}. (Best viewed in color) The environment and tasks for evaluation.
  \textbf{(a)} BabyAI Trv-1 task with single-side door obstacle;
  \textbf{(b)} BabyAI Trv-2 task with double-side door obstacles;
  \textbf{(c)} BabyAI Numerical Q\&A task;
  \textbf{(d)} Two VirtualHome household environments (left: VH-1; right: VH-2) and an examplar task.
  }
  \vspace*{-0.15in}
 \label{fig:expSettings}
\end{figure*}
%%%%%%%%%%%%%%%%%%%%%%%%%%%%%%%%%%%%%%%%%%%%%%%%%%%%%%%%%%%%%

\section{Experimental Settings}

% \Ben{I think it would be good to add a sentence here about the fact that we are interested in both global (i.e. Q & A) and local tasks (i.e. planning). The idea is that we can solve either. This doesn't have to be rigorous, just a way to keep a theme going through the paper.}
We evaluate our methods on a series of numerical Q\&A (NumQ\&A) and planning tasks, which require both global and local spatial reasoning, in the BabyAI \citep{babyai, minigrid} and VirtualHome (VH) \citep{virtualhome} environments. For each environment, we provide an unified scene graph schema consistent across epoches with distinct scene graphs. Each task requires reasoning on both the spatial structure and the semantic information encoded in the graph. For evaluation metric, we use the \textbf{success rate}, defined as the ratio of the trials where the task solving is successful. The success is defined as either providing the correct answer for the Q\&A tasks or achieving the desired outcome for the planning tasks.
Note that all experiments in this paper are conducted in the static setting, where the tested methods generate solutions solely based on the initial scene graph without interacting with the environment or modifying the graph. We also provide preliminary results under the dynamic settings in the supplementary.

Unless otherwise specified, we use GPT-4o as the backbone LLM for all methods. We implement \RwR with AutoGen \citep{autogen}. For all methods, we set both temperature and random seed to 0.

\paragraph{Baselines} Following NLGraph \citep{NLGraph}, we compare our approach with several direct graph prompting methods including: \textbf{zero-shot prompting} (\textsc{zero-shot}), \textbf{Zero-Shot Chain-of-Thought} (\textsc{0-cot}) \citep{zeroShotCot}, \textbf{Least-to-Most} (\textsc{ltm}) \citep{LTM}, \textbf{Chain-of-Thought} (\textsc{cot}) \citep{CoT}, \textbf{Build-a-Graph} (\textsc{bag}) \citep{NLGraph}, \textbf{Algorithmic Prompting} (\textsc{algorithm}) \citep{NLGraph}. In addition to the few-shot examples, \textsc{algorithm} also requires a language description of the task solving method. 
We also compare against \textbf{SayPlan} \citep{sayplan}, a retrieve-then-reason baseline specifically designed for the scene graphs, and \textbf{ReAct} \citep{react}, a generic iterative reasoning and acting approach that invokes database APIs to aggregate information. 
Since SayPlan does not release the source code, we evaluate with our implementation of the method.
For ReAct, we curate a graph traversal tool
\texttt{\small expand(nodeID)}
to retrieve attributes of a specified node plus the IDs and attributes of all its neighbor nodes.  
We also provide it with any reasoning tools available to \RwR depending on the task.
We annotate few-shot examples involving detailed task solving process for both SayPlan and ReAct following their format.

\begin{figure}[t!]
    \centering
    \vspace*{-0pt}
    \scalebox{0.67}{
    	\begin{tikzpicture}
         \node[anchor=north west] at (0in,0in)
          {{\includegraphics[width=1.0\linewidth,clip=true,trim=0
          260pt 580pt 15pt]{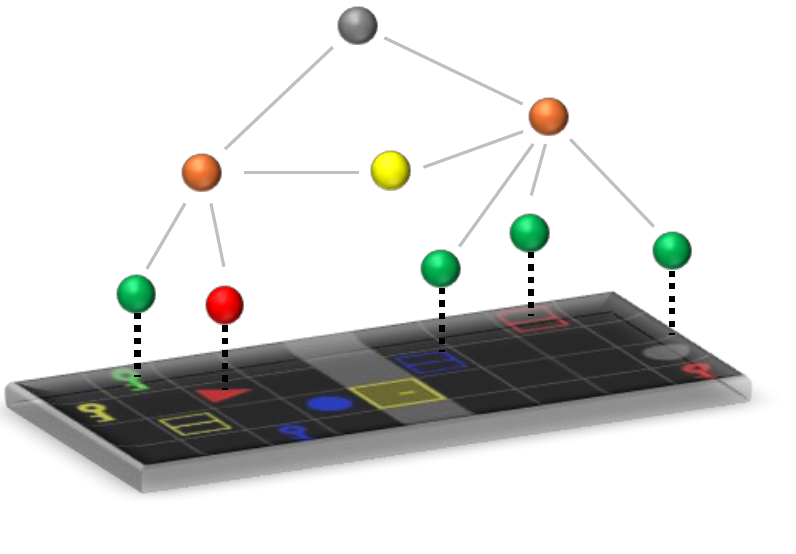}}};
        \end{tikzpicture}
    }
    \vspace*{-15pt}
    \caption{
        \textbf{BabyAI Scene Graph Representation}. Graph nodes represent \green{items}, \red{agents}, \orange{rooms}, and \yellow{doors}. Edges indicate items or agents located inside a  room, or doors that connect rooms. Room nodes are connected to a \gray{root} node.
    }
    \label{fig:BabyAISG}
    \vspace*{-10pt}
\end{figure}

Next few subsections summarize the design of the environments and tasks. More details on the scene graph design as well as the schema are shown in the supplementary.

\subsection{2D Grid World Numerical Q\&A}
Our first experiment is on a numerical Q\&A task in a customized 9-room 2D BabyAI \citep{babyai} environment, as shown in Figure \ref{fig:expSettings}(c).
We generate scene graph representation of the environment following the hierachical graph design from 3DSG \citep{3dsg}, illustrated in Figure \ref{fig:BabyAISG}. Specifically, the graph represents the spatial scene layout through three levels: root, rooms, and objects, with additional door nodes connecting room pairs.
 
Following SayPlan \citep{sayplan}, we design the following question template: \texttt{\small find the color of the \string{TARGET\_OBJECT\string} in a room next to the room with \string{NUM\_IDENTIFIER\string} \string{COLOR\_IDENTIFIER\string} \string{IDENTIFIER\_OBJECT\string}}, where contents in curley brackets are populated based on each environment instance. 
The environment and question pairs are designed to ensure that there is only one answer.

We test each method in 100 task instances. We manually annotate the few-shot demonstrations for few-shot methods. We also annotate the task solving process for SayPlan, ReAct, and \RwR following their respective response formats.

\subsection{2D Grid World Traversal Planning}
\label{sec:BabyAITrv}
We also test on the traversal planning in BabyAI, where the task is to generate a sequence of node-centric actions to pick up a target item. We design three atomic actions, including (1) \texttt{\small pickup(nodeID)}: Walk to and pickup an object by the node ID; (2) \texttt{\small remove(nodeID)}: Walk to and remove an object by the node ID; (3) \texttt{\small open(nodeID)}: Walk to and open a door by the node ID. 
%We directly query \RwR and all baselines to generate the actions in the format above.

As shown in Figure \ref{fig:expSettings}(a)(b), the traversal planning task is tested in two related double-room environments, both of which require the agent to pick up the key of the correct color to unlock the door, remove any obstacle that blocks the door, open the door, and pick up the target. The difference is that the first environment, dubbed \textbf{Trv\-1}, contains only the agent-side obstacle, whereas the second environment, dubbed \textbf{Trv\-2}, contains another target-side obstacle. We generate the in-context examples \textit{only in Trv\-1} , and test if the methods can extrapolate to Trv\-2. As before, we evaluate each method in 100 times in different instance of both types of the environment. 
For \RwR and ReAct, we provide the reasoning tool \texttt{\small traversal\_room} programmed based on the $A^*$ algorithm, which identifies the item to remove in order to reach from an initial to a desired location within the same room.

\subsection{Household Task Planning}
The last evaluation is in two VirtualHome (VH) \citep{virtualhome} environments shown in Figure \ref{fig:expSettings}(d), denoted as \textbf{VH-1} and \textbf{VH-2}, respectively. We use the built-in environmental graph as the scene graph. Compared to BabyAI, VH environments have larger state space and action space, containing 115 object instances, 8 relationship types, and multiple object properties and states. Hence, VH environments are more challenging with richer information in the graphs.
% The action space is: $\mathcal{A}$ = \texttt{\{\}}. 
For each environment, we adopt the 10 household tasks from ProgPrompt \citep{progprompt}, such as \texttt{\small "put the soap in the bathroom cabinet"}, and query each method for the action sequence in the VH action format to accomplish the task. 
% As before, we task each method to directly generate the plan in the VH action format. It includes \texttt{\small [action\_name]<object\_name>(object\_id)} for one argument actions, and \texttt{\small [action\_name]<object\_name1>(object\_id1)<object\_name2>(object\_id2)} for two argument actions.
We use two of the tasks, together with the ground truth actions, as the few-shot examples, and test with the other eight. We follow CoELA \cite{coopEmbod} to specify the task as the desired states. For example, the task of above is specified as \texttt{\small soap INSIDE bathroomcabinet}. 
% To achieve the desired state, LLMs need to reason over the current state of the environment in order to identify the sequence of actions that ultimately achieve the achieve the desired outcome. A plan is considered successful if the desired states are reached after simulation. 
% For more details, please refer to Appendix \ref{app:VHDetail}.
Due to the unavailability of reasoning process annotation behind the solution, we do not include it in the few-shot prompt for \RwR, and do not test ReAct and SayPlan as they do not work well without the demonstrations.

%% BabyAI in one table
% \begin{table}[t!]
\begin{table*}[t!]
    \centering
    \setlength\tabcolsep{6.pt}
    \setlength\extrarowheight{-9pt}
    % \begin{tabular}{l c c c c c c c c c c c c}
    \begin{tabular}{l c c c c c c c c c c}
        \toprule[1.5pt]
              & \multicolumn{3}{c}{Zero-Shot} & \multicolumn{6}{c}{Few-Shot}
             \\
             \cmidrule(lr){2-4}
             \cmidrule(lr){5-10}
             \\
             \textbf{Task} & ZeroShot & 0-CoT & LTM
             %& \textbf{\RwR} 
             & CoT & BAG & Alg & SayPlan & ReAct
             % & \makecell{\textbf{\RwR} \\(FS)} & \makecell{\textbf{\RwR} \\ (Alg)}
             & \RwR
             \\
        \midrule[1pt]
             NumQ\&A & 29\% & 60\% & 52\% &  56\% & 49\% & 67\% & 
             35\% &
             86\% & 98\% 
             \\
             Trv-1 & 20\% & 50\% & 63\% & 54\% & 71\% & 45\% & 18\% & 
             94\% & 97\% 
             \\
             Trv-2 & 13\% & 16\% & 20\% & 0\% & 0\% & 11\% & 0\% & 
             95\% & 96\% 
             \\
         \bottomrule[1.5pt]
    \end{tabular}
    \caption{\textbf{Results in BabyAI}.
    \RwR outperforms all baseline methods, showing efficacy of our design.
    % \RwR achieves the best performance across all tasks in both zero-shot and few-shot settings, showing that \RwR (1) is effective in solving spatial tasks; (2) can harness the information from in-context examples and extrapolate better to unseen tasks. We highlight the top-1 performance under the zero-shot setting and top-2 performances, including ties, under the few-shot setting.
    }
    \label{tab:BabyAI}
    % \vspace{-0.2in}
\end{table*}

\begin{table}[t!]
    \centering
    \setlength\tabcolsep{12.pt}
    \renewcommand{\arraystretch}{1.2}
	\begin{tabular}{l  c  c }
        \toprule[1.5pt]
        Method 
        % & \makecell{Few-Shot \\ Examples} 
        & VH-1 & VH-2 \\
        \hline 
         \textbf{ZeroShot}  & 7/8 & 6/8 \\
         \textbf{0-CoT}     & 7/8 & 6/8 \\
         \textbf{LTM}       & 7/8 & 5/8 \\
         \hline
         \textbf{CoT}       
                        % & \bluecheck     
                            & 7/8 & 6/8 \\
         \textbf{BAG}       
                            % & \bluecheck 
                            & 7/8 & 5/8 \\
         \hline
         \textbf{\RwR} & \bf{8/8} & \bf{8/8} \\
         % \cmidrule(r){1-2} \cmidrule(lr){3-14} \cmidrule{15-16}
        \bottomrule[1.5pt]
    \end{tabular}
    \caption{
        \textbf{Results in VirtualHome}. The number of accompolished tasks out of 8.
        The superior performance of \RwR shows its practicality in realistic environments.
    }\label{tab:VHResults}
    \vspace{-12pt}
\end{table}

\section{Results and Analysis}

\subsection{Experiment Results}
% We also present qualitative results in Appendix~\ref{app:RwRTrvDemo} (BabyAI), Appendix~\ref{app:RwRVHDemo} (VirtualHome), and Appendix~\ref{app:BaselineFailures} (Baseline failure cases). We provide analysis on the compute cost with the iterative design in Appendix~\ref{app:ComputeAnly}.
% This section shows qualitative results. 
%We provide qualitative results in the supplementary.

\paragraph{Numerical Q\&A Resutls} The results are tabulated in Table \ref{tab:BabyAI}. 
For the baselines, few-shot methods CoT underperforms even compared to the zero-shot counterpart 0-CoT, suggesting that few-shot prompts do not show consistent effect in this task.
This is due to the fact that although establishing reasoning trace for this task is simple, solving atomic spatial tasks such as counting the item or locating the neighboring rooms is not straightforward for LLMs when processing large graphs as text.
On the other hand, ReAct outperforms other graph prompting baselines by at least 19\%, showing the effectiveness of Reason-while-Retrieve strategy in addressing the aforementioned issue. 
However, ReAct is still limited by the API-based information retrieval, requiring multiple calls for simple sub-problem such as "finding a room with 4 green balls". 
In contrast, the program-based information retrieval in our method is able to solve the sub-problem, and the multi-agent framework ensures that the reasoning is not misdirected by the long program generated by the Retriever. Both factors attributes to the 12\% performance advantage of our method \RwR agains ReAct.

\paragraph{2D Traversal Results} Table \ref{tab:BabyAI} also reports the success rate in the traversal task. 
The few-shot prompts demonstrate more advantage in this task, showing as CoT outperforms 0-CoT by $4\%$ and BAG achieving the best performance compared to other graph prompting methods under the in-domain Trv-1.
However, the advantage does not extrapolate with slight domain change. In Trv-2, those few-shot methods all under-perform compared to zero-shot methods, with CoT, BAG, and SayPlan even dropping to $0\%$. This again suggest that graph prompting is not an effective solution for scene graph reasoning.
On the other hand, ReAct and our method again show better capacity, achieving more than $20\%$ lead in success rate compared to other methods.
While our method still outperforms ReAct, the gap is small, with only $1\%$ or $3\%$. 

\paragraph{Household Task Planning Results}
The planning success rate on the 8 tasks in the 2 VH environments are shown in Table \ref{tab:VHResults}. We observe that all baselines consistently fail to address the precondition of the planned action. For example, all of them failed to generate \texttt{\small [open] <garbagecan> (ID)} before \texttt{\small [putin] <plum> (ID) <garbagecan> (ID)}, forgetting that the state of the garbage can is \texttt{\small state:\{CLOSED\}} from the extensive graph input. On the other hand, \RwR doesn't process the entire graph. Instead, it queries for the specific object information, which helps to better determine the action parameter and examine the action preconditions. 
% For qualitatitve demonstration, please refer to Figure \ref{fig:vhQual} for an examplar task and solution by our method.
% \Ben{No comparison to ReAct on this data? We should say why not.}

\subsection{\RwR v.s. ReAct}
\label{sec:vsReAct}
As explained before, despite the same iterative Reason-while-Retrieve strategy, our method differs from ReAct in program-based graph interaction, schema contextualization, and the multi-agent design that separates reason and retrieve contexts.
In this section, we further justify our designs through ablation on both \RwR and ReAct by testing the following variants:
\begin{itemize}
    \item \textbf{ReAct-limit}: 
    This variant is the ReAct method with weaker graph traversal APIs.
    We removes the \texttt{\small expand(nodeID)} API and breaks it down into
    \texttt{\small get\_neighbors(nodeID)} and \texttt{\small get\_attrs(nodeID)},
    which obtains only the IDs of neighbor nodes and attributes of a specific node, respectively.
    With this modification, each function obtains less information compared to before, requiring more API calls to aggregate information for a reasoning substep.
    This variant examines ReAct's sensitivity to the API capacity, which impacts its compatibility to different graph databases. 
    Takes our case for example, the \texttt{\small get\_neighbors} and \texttt{\small get\_attrs} functionality are directly provided by the NetworkX library, whereas \texttt{\small expand} requires manual curation.

    \item \textbf{\RwR-limit}: 
    To verify the efficacy of the Reason-Retrieve separation with multi-agent, 
    we evaluate this variant of our method without programming-based information retrieval. 
    Instead, the Retriever is re-designed as a \textit{ReAct-limit} agent, which only relies on tool calling with APIs of limited capacity. 
    Unlike ReAct-limit, the contexts of reasoning and retrieval are not shared.

\end{itemize}
    
All variants are tested in BabyAI tasks. 

\paragraph{Results} The results are collected in Table \ref{tab:vsReAct}. 
Compared to ReAct, the performance of ReAct-limit drops significantly. 
The success rates of all three tasks are lowered by more than $36\%$ solely due to the breakdown of the API function.
This result verifies the drawback of ReAct, which is its over-reliance on the API quality. 
In the case where the information gathered from APIs has larger semantic gap to the reasoning demand, increased number of API calls are necessary. 
Without separating the reasoning and retrieval history, the context for both stages build up with more iteration steps, leading to higher chance of hallucination.
On the other hand, even with the same set of graph APIs, \RwR-limit still outperforms ReAct-limit on all three tasks, with more $35\%$ gap on both traversal tasks.
This validates the importance of distributing reasoning and retrieval to multiple agents, which effectively filter the context based on the functionality of each component.

% ==========================================
\subsection{Small Language Models Performance}
We conduct studies on the choice of LLM backbone, especially of the open-source Small Language Models (SLMs).
Specifically, we test \RwR, together with ZeroShot, 0-CoT, LTM, CoT, BAG, Alg, and ReAct baselines, with Phi4-14B~\citep{phi4}, Qwen3-14B~\citep{qwen3}, and DeepSeek-7B~\cite{deepseek} models on the BabyAI NumQ\&A task. Each method is tested with 20 trials and the success rate is collected.

\begin{figure}[t!]
    \centering
    \vspace*{-0pt}
    \scalebox{0.93}{
    	\begin{tikzpicture}
         \node[anchor=north west] at (0in,0in)
          {{\includegraphics[width=1.0\linewidth,clip=true,trim=0
          210pt 380pt 0pt]{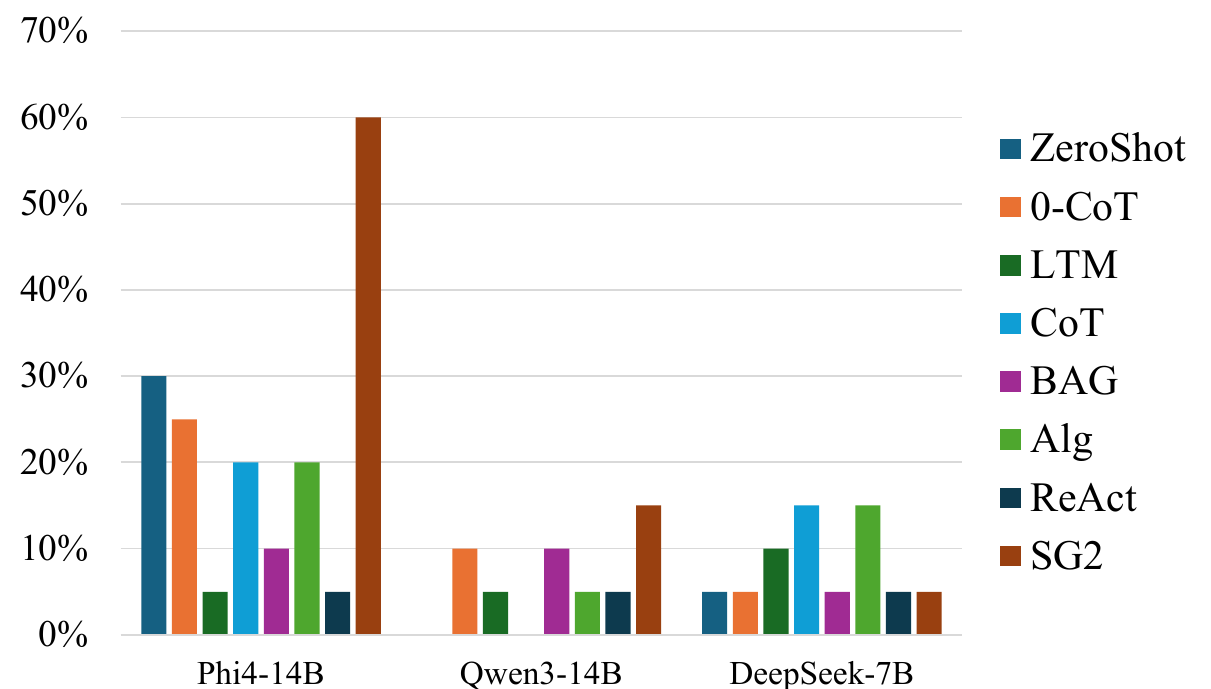}}};
        \end{tikzpicture}
    }
    % \vspace*{-15pt}
    \caption{
        \textbf{Small Language Models (SLMs) results on the BabyAI NumQ\&A task}.
    }
    \label{fig:SLMsResults}
    % \vspace*{-10pt}
\end{figure}

\paragraph{Results}
The results are illustrated in the Figure.~\ref{fig:SLMsResults}. 
The performance of all baseline models drop significantly with SLMs, with the best success rate being only $30\%$ with the Phi4-14B model and less than $20\%$ with the Qwen3-14B or deepseek-7B, suggesting the weak ability of SLMs to comprehend graph structure with textual inputs.
On the other hand, despite the equally poor performance of our method with Qwen3-14B or deepseek-7B, \RwR achieves $60\%$ success rate with Phi-14B, which doubles compared to even the best-performing baseline.
This suggests that reasoning abstractly with graph schema might be a simpler compared to comprehending the entire textualized graph for SLMs, 
which shows the potential of utilizing multi-agent and code-writing in graph reasoning tasks.

\begin{table}[t!]
    \centering
    \renewcommand{\arraystretch}{1.2}
    \begin{tabular}{l c c c}
        \toprule[1.5pt]
        Method  & Num Q\&A & Trv-1 & Trv-2\\
         \hline
         \textbf{ReAct-limit} & 40\% & 58\% & 11\%\\
         \textbf{ReAct}  & 86\% & 94\% & 95\%\\
         \textbf{\RwR-limit} &  47\% & 93\% & 70\%\\
         \textbf{\RwR} & 98\% & 97\% &  96\% \\
         % \cmidrule(r){1-2} \cmidrule(lr){3-14} \cmidrule{15-16}
        \bottomrule[1.5pt]
    \end{tabular}
    \caption{
        \textbf{ReAct and \RwR Comparison}. 
        \RwR-limit outperforms ReAct-limit, showing the efficacy of multi-agent system.
        \RwR further improves the results and achieves the best performance on all three tasks, validating the design of graph processing with schema-guided abstract programs.
    }\label{tab:vsReAct}
    % \vspace{-10pt}
\end{table}

\section{Conclusion}
% \Ben{In the conclusion, I'd really like to draw out the point that our approach is just a variation on the very common theme in computing: Separation of logic from the data. The Retriever-Reason split does this for the reasoner. And the Schema does this for the Retriever. This is a very powerful message, as it ties the whole idea into something that anyone who has worked with computers will have thought about before.}
In this work, we propose \RwR, a schema-guided, multi-agent framework performs iterative reason-while-retrive on scene graphs.
The use of schema induces the idea of separating logic from the data, allowing the Reasoner to perform task planning abstractly, and the Retriever to write symbolic code for accessing necessary data while filtering irrelevent information. 
The multi-agent design facilitates the realization of the above process, by separating the reason and retrieve process along with their contextual inputs as well as improving the robustness of the pipeline. 
Our experiments show that our method achieves the best results on all tested benchmarks. What's more, by conducting ablation study on our method and ReAct, we verify the efficacy of both the schema-guided programming as well as the multi-agent designs.

Future work could explore the flexibility of \RwR framework to seamlessly integrate  additional agents with new specialties. Potential new agents involve a verifier agent to correct the solution using graph information and new modality agent to process richer information. Reasoning trace optimization could also be explored, as the conversation rounds scale with task difficulty and agent numbers.

\bibliography{reference}

% % NOTE: comment this out for arxiv version
% % Check whether the conference requires a reproducibility checklist to be included in the paper.
% % If so, you can uncomment the following line and ajust the path to include it.
% \newpage
% \input{AAAI2026/ReproducibilityChecklist}

\clearpage

%%%%%%%%%%%%%%%%%%%%%%%%%%%%%%%%%%%%%%%%%%%%%%%%%%%%%%%%%%%%%
%%  Prompts
%%%%%%%%%%%%%%%%%%%%%%%%%%%%%%%%%%%%%%%%%%%%%%%%%%%%%%%%%%%%%
\section{Prompt Templates for \RwR}
\label{app:PromptTemplate}
\RwR adopts template-based prompt generation for all agents. The templates are shown in the tables below, where the colored placeholders are replaced with environment-specific information to obtain the prompts.

\begin{center}
\captionof{table}{Retriever \RetCoder Template}
\label{tab:RetCoderPrompt}
\begin{supertabular}{|p{0.94\columnwidth}|}
\hline
\textbf{Retriever \RetCoder Prompt Template} \\
\hline
\begin{lstlisting}[style=prompt]
Given the description of scene graph schema and scene information, please write python code to address information retrieval queries.

(*@\textcolor{cyan}{\textbf{[ENVIRONMENT EXPLANATION]}}@*)
(*@\textcolor{cyan}{\textbf{[SCENE GRAPH SCHEMA]}}@*)

Given each query, please write python code to retrieve the queried information from the graph. 
Please respond only with python code. Please wrap the code in a python code block:```python [YOUR CODE]```

Please print out the queried node(s) and/or link(s) in your code.
If the execution error is received, please re-write the code.
\end{lstlisting}
\\
\hline
\end{supertabular}
\end{center}

\begin{center}
\captionof{table}{Retriever \RetLan Template}
\label{tab:RetLanPrompt}
\begin{supertabular}{|p{0.94\columnwidth}|}
\hline
\textbf{Retriever \RetLan Prompt Template} \\
\hline
\begin{lstlisting}[style=prompt]
Given a information retrieval query and a code execution result, determine if the execution result addresses the query. If the query is addressed or partially answered, please only summarize the result. 

If the query is not addressed, please only respond 'NOT ADDRESSED'.
\end{lstlisting}
\\
\hline
\end{supertabular}
\end{center}

\begin{center}
\captionof{table}{Reasoner \ReaCoder Template}
\label{tab:ReaCoderPrompt}
\begin{supertabular}{|p{0.94\columnwidth}|}
\hline
\textbf{Reasoner \ReaCoder Prompt Template} \\
\hline
\begin{lstlisting}[style=prompt]
You are provided with a list of python tool functions and their annotations. 
Given a messagae containing the tool to use and arguments, please convert them to executable python code. The list of tools and their annotations:

(*@\textcolor{cyan}{\textbf{[TOOL ANNOTATIONs]}}@*)

Please respond with only the python code wrapped in a python code block:```python [YOUR CODE]```.
In your code, import the tool via 'from functions import TOOL_FUNCTION'.
Print out the result in the code.
\end{lstlisting}
\\
\hline
\end{supertabular}
\end{center}

\begin{center}
\captionof{table}{Reasoner \ReaLan Template}
\label{tab:ReaLanPrompt}
\begin{supertabular}{|p{0.94\columnwidth}|}
\hline
\textbf{Reasoner \ReaLan Prompt Template} \\
\hline
\begin{lstlisting}[style=prompt]
Given the description of scene graph schema, scene information, and potentially the action space of an agent in the scene, 
please collaborate with a Retriever and a tool executor in natural language to solve given tasks.
You don't know about any graph information. Please always query the retriever for any information you need.
(*@\textcolor{cyan}{\textbf{[ENVIRONMENT EXPLANATION]}}@*)
(*@\textcolor{cyan}{\textbf{[SCENE GRAPH SCHEMA]}}@*)
(*@\textcolor{cyan}{\textbf{[ACTION SPACE]}}@*) (for planning tasks)

[Response Modes and Formats]
You can only respond in three modes in natural language:
1. Query mode. Reason on what information do you need to solve the task, and query for the information from the retriever.
2. Tool calling mode. Call a function from the given function set to address a substep.
3. Solution mode. Give the solution to the task.

(*@\textcolor{cyan}{\textbf{[TOOL ANNOTATIONs]}}@*)

Please always format all your responses as follows:

    [Explanation]
    Explain your reasoning process succinctly.

    [Mode]
    Only one of QUERY or SOLUTION or TOOL.
    Note that the SOLUTION mode will terminate the conversation. So only use it when you have the entire solution.

    [Content]
    If QUERY, then give the query here in 1 sentence. Be explicit about what attributes you need based on the schema.
    If TOOL, then give the function to call. Clearly list out the function name and all the argument values. All argument values must be queried from the scene graph.
    If SOLUTION, then give the solution here.

(*@\textcolor{cyan}{\textbf{[IN-CONTEXT EXAMPLES]}}@*)
\end{lstlisting}
\\
\hline
\end{supertabular}
\end{center}

\pagebreak

%%%%%%%%%%%%%%%%%%%%%%%%%%%%%%%%%%%%%%%%%%%%%%%%%%%%%%%%%%%%%
%%  Environment Details
%%%%%%%%%%%%%%%%%%%%%%%%%%%%%%%%%%%%%%%%%%%%%%%%%%%%%%%%%%%%%

% \begin{table*}[!htp]
%     \centering
%     % \setlength\tabcolsep{3.2pt}
%     \begin{tabular}{r l}
%         \toprule[1.5pt]
%               \textbf{Task Name} & \textbf{State Specification }
%               \\
%               \hline
%              Watch TV & tv ON \\
%              Turn off tablelamp & tablelamp OFF \\
%              put the soap in the bathroomcabinet & barsoap INSDIE bathroomcabinet \\
%              throw away plum & plum INSIDE garbagecan \\
%              make toast & breadslice INSIDE toaster; breadslice HEATED \\
%          \bottomrule[1.5pt]
%     \end{tabular}
%     \caption{\textbf State-based Task Specification in VirtualHome}
%     \label{tab:VHEgTasks}
% \end{table*}
% % \vspace{1em}   

\twocolumn[  % ← Insert full-width content before 2-column resumes

\begin{center}
\label{tab:VHEgTasks}
\begin{tabular}{rl}
\hline
\textbf{Task Name} & \textbf{State Specification} \\
\hline
WatchTV & tv ON \\
Turn off tablelamp & tablelamp OFF \\
put the soap in the bathroomcabinet & barsoap INSIDE bathroomcabinet \\
throw away plum & plum INSIDE garbagecan \\
make toast & breadslice INSIDE toaster, breadslice HEATED \\
\hline
\end{tabular}
\captionof{table}{State-based Task Specification in VirtualHome}
\end{center}

]  % ← End of full-width content

\section{Environment Details}
\label{app:EnvDetail}

\subsection{BabyAI Environment and Scene Graph Details}
\label{app:babyAIDetail}
\paragraph{Node attributes} The node attributes in BabyAI scene graph involve:
\begin{itemize}
    \item \textbf{"type"}: String. The type of the element type. Choices: \[root, room, agent, key, door, box, ball\]
    \item \textbf{"color"}: String. For doors and items. The color of the element.
    \item \textbf{"coordinate“}: List of integer. Exist for all types of nodes except for the root node. For room nodes, the top left corner coordinate. For other nodes, the 2D coordinate in the grid.
    \item \textbf{"is\_locked"}: Binary. For door. State indicating if a door is locked or not.
    \item \textbf{"size"}: List of integer. For room. The size of a room. 
\end{itemize}

\subsection{VirtualHome Environment and Scene Graph Details}\label{app:VHDetail}
VirtualHome is a large indoor environments with more extensive graph node and edge space as well as agent action space. 
\paragraph{Node attributes} in VirtualHome involve the following types:
\texttt{\small 'id', 'category', 'class\_name', 'prefab\_name', 'obj\_transform', 'bounding\_box', 'properties', 'states'}
% \begin{itemize}
%     \item \textbf{'id'}: Int. Node id.
%     \item \textbf{'category'}: Str. Meta category. E.g. "Room".
%     \item \textbf{'class\_name'}: Str. Specific class name. E.g. "bathroom". 
%     \item \textbf{'prefab\_name'}: Str. Instance name.
%     \item \textbf{'obj\_transform'}: Dict. {'position': 3D vector, 'rotation': Quaternion form as 4D vector, 'scale': 3D vector} 
%     \item \textbf{'bounding\_box'}: Dict. {'center': 3D vector, "size": 3D vector} 
%     \item \textbf{'properties'}: List. Object properties. Determine the action that can act upon it. 
%     \item \textbf{'states'}: List. Object states. Full list of available states: ['CLOSED', 'OPEN', 'ON', 'OFF', 'SITTING', 'DIRTY', 'CLEAN', 'LYING', 'PLUGGED\_IN', 'PLUGGED\_OUT', 'HEATED', 'WASHED']
% \end{itemize}

\paragraph{Edge attributes} in VirtualHome include the following categories:
\texttt{\small 'from\_id', 'to\_id', 'relationships'},
where the types of \texttt{\small 'relationships'} are:
\texttt{\small 'ON', 'INSIDE', 'BETWEEN', 'CLOSE', 'FACING', 'HOLDS\_RH', 'HOLD\_LH', 'SITTING'},
% \begin{itemize}
%     \item \textbf{'from\_id'}: Int. Id of node in the from relationship.
%     \item \textbf{'to\_id'}: Int. Id of node in the to relationship.
%     \item \textbf{'relationships'}: Str. Relationship between the 2 objects. Available relationships:
%     \begin{itemize}
%         \item \textbf{'ON'}: Object from\_id is on top of object to\_id.
%         \item \textbf{'INSIDE'}: Object from\_id is inside of object to\_id.
%         \item \textbf{'BETWEEN'}: Used for doors. Door connects with room to\_id.
%         \item \textbf{'CLOSE'}: Object from\_id is close to object to\_id (< 1.5 metres).
%         \item \textbf{'FACING'}: Object to\_id is visible from objects from\_id and distance is < 5 metres. If object1 is a sofa or a chair it should also be turned towards object2.
%         \item \textbf{'HOLDS\_RH'}: Character from\_id holds object to\_id with the right hand.
%         \item \textbf{'HOLD\_LH'}: Character from\_id holds object to\_id with the left hand.
%         \item \textbf{'SITTING'}: Character from\_id is sitting in object to\_id.
%     \end{itemize}
% \end{itemize}

\paragraph{Action Space} in VirtualHome encompasses:
\begin{itemize}
    \item \textbf{[walk] <class\_name> (id)}: Walk to an object. 
    \item \textbf{[grab] <class\_name> (id)}: Grab an object. Requires that the agent has walked to that object first.
    \item \textbf{[open] <class\_name> (id)}: Open an object. Requires that the agent has walked to that object first.
    \item \textbf{[close] <class\_name> (id)}: Close an object. Requires that the agent has walked to that object first.
    \item \textbf{[switchon] <class\_name> (id)}: Turn an object on. Requires that the agent has walked to that object first.
    \item \textbf{[switchoff] <class\_name> (id)}: Turn an object off. Requires that the agent has walked to that object first.
    \item \textbf{[sit] <class\_name> (id)}: Sit on an object. Requires that the agent has walked to that object first.
    \item \textbf{[putin] <class\_name1> (id1) <class\_name2> (id1)}: Put object 1 inside object 2. Requires that the agent is holding object 1 and has walked to the object 2.
    \item \textbf{[putback] <class\_name1> (id1) <class\_name2> (id1)}: Put object 1 on object 2. Requires that the agent is holding object 1 and has walked to the object 2.
\end{itemize}

\paragraph{Example Task and State-based Specifications in VH-1}
We show the 5 example tasks and their desired final state in the VH-1 environment in Table \ref{tab:VHEgTasks}.

\section{Qualitative results}

%%%%% figure %%%%%%%%%%%%%%%%%%%%%%%%%%%%%%%%%%%%%%%%%%%%%%%%%
\begin{figure*}[th!]
 
  % \vspace*{-0.1in}
  \centering
  \scalebox{0.92}{
    \begin{tikzpicture}
     \node[anchor=north west] at (0in,0in)
      {{\includegraphics[width=1.0\linewidth,clip=true,trim=0
      250pt 0pt 0]{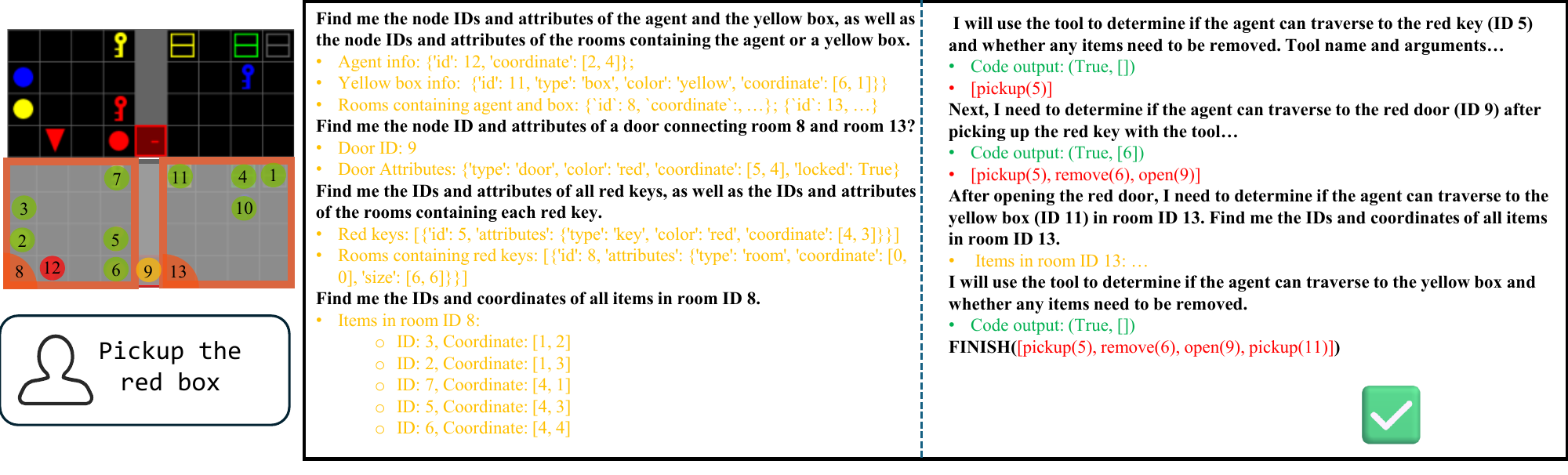}}};
    \end{tikzpicture}
  }
  \caption{\textbf{Example \RwR Traversal task solving process (\ReaLan point-of-view)}.  It shows the queries or analysis generated by the \ReaLan (in black), information obtained from the Retriever (in \yellow{yellow}), the results of tool execution that processes obtained information (in \green{green}), and the derived plan (in \red{red}). The final plan can successfully achieve the mission shown on the left.
  }
  \vspace*{-0.1in}
 \label{fig:rwrTrvDemo}
\end{figure*}
\begin{figure*}[t!]
 
  % \vspace*{-0.1in}
  \centering
  \scalebox{0.92}{
    \begin{tikzpicture}
     \node[anchor=north west] at (0in,0in)
      {{\includegraphics[width=1.0\linewidth,clip=true,trim=0
      180pt 00pt 0]{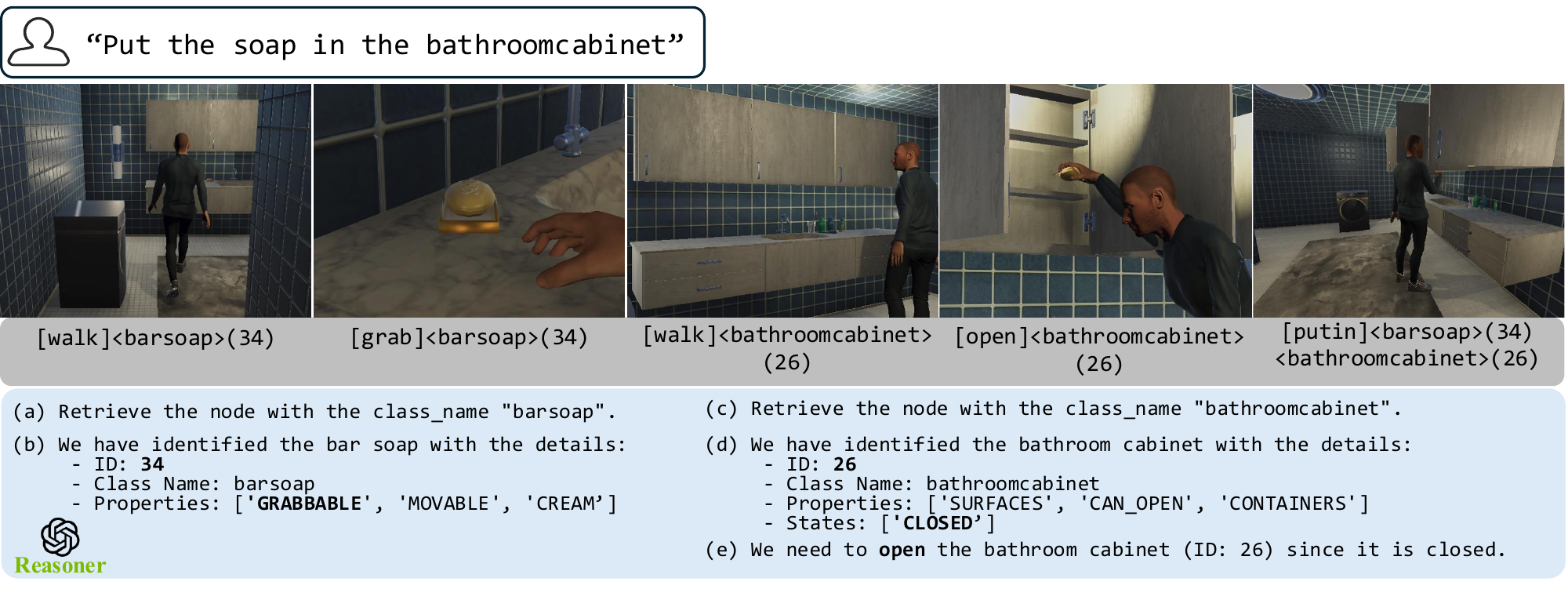}}};
    \end{tikzpicture}
  }
  \caption{\textbf{VirtualHome Qualitative Demonstration}. 
  \textbf{Top row}: Plan Execution;
  \textbf{Middle row}: Generated plan in the VirtualHome action format.
  \textbf{Bottom row}: \RwR Snippet of the Reasoner-side generation leading to the plan. 
  }
  \vspace*{-0.1in}
 \label{fig:vhQual}
\end{figure*}
%%%%%%%%%%%%%%%%%%%%%%%%%%%%%%%%%%%%%%%%%%%%%%%%%%%%%%%%%%%%%

\subsection{Qualitative Results in BabyAI Traversal Task}
\label{app:RwRTrvDemo}
We qualitatively demonstrate how \RwR addresses a challenging BabyAI traversal task in Figure \ref{fig:rwrTrvDemo}. It shows the task solving process from the \ReaLan's perspective --- including its thought process, the output information query or tool-calling request, and the corresponding graph data or tool execution results received. It clearly demonstrates that \RwR is capable of grounding the plan to the environment by iteratively retrieving graph information based on the task solving process and establishing the next step towards solution based on the past retrieved information.

\subsection{Qualitative results in VirtualHome}
\label{app:RwRVHDemo}
The exemplar result in VirtualHome is shown in Figure~\ref{fig:vhQual}. We only show core outputs from the Reasoner module. The result shows that our method is able to generate the correct reasoning trace solely based on the graph schema, raise corresponding queries, and use the returned information to generate the correct plan for a given task.

\begin{figure*}[ht!]
    \centering
    \scalebox{0.85}{
    % First subfigure
    \begin{subfigure}{0.48\linewidth}
        \centering
        \includegraphics[width=\linewidth]{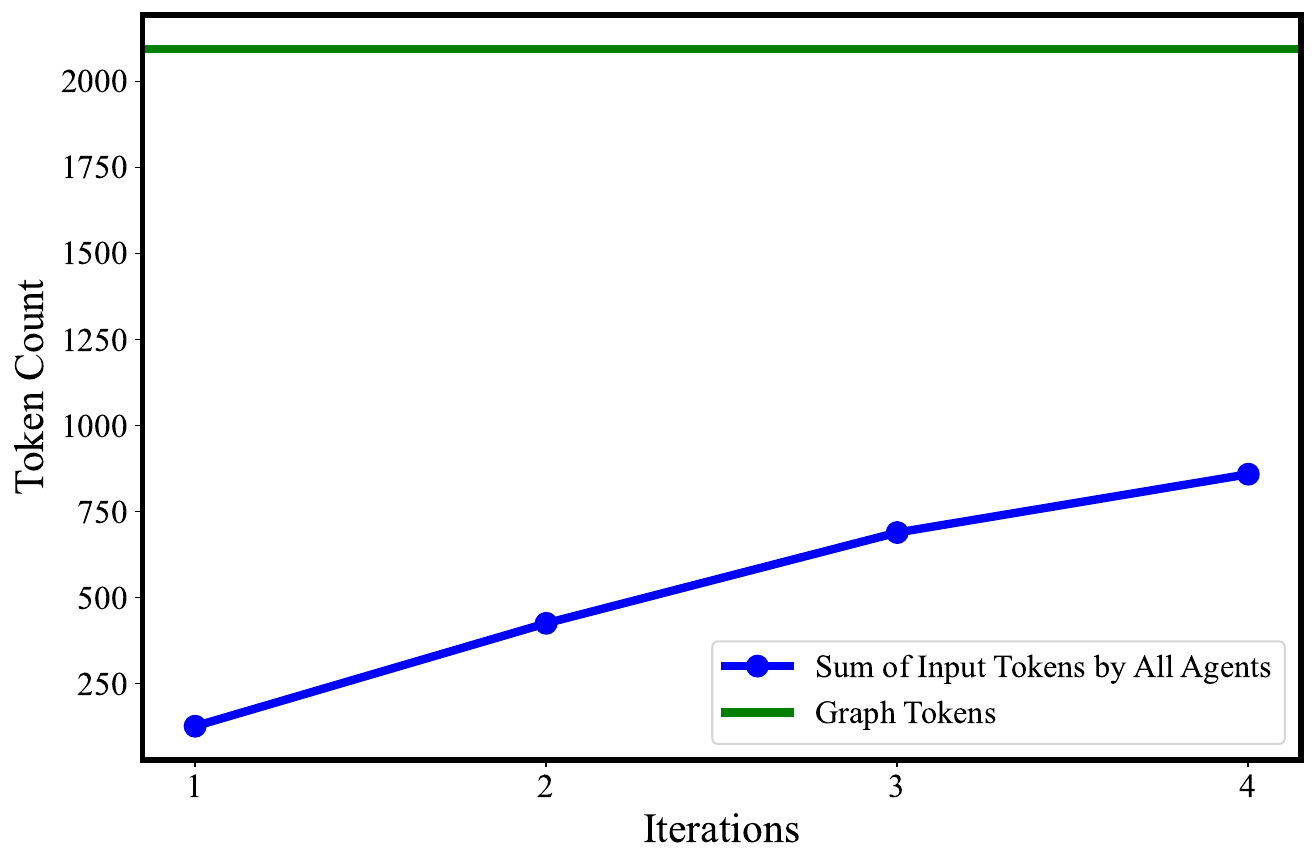}
        \caption{NumQ\&A}
        \label{fig:CompNumQA}
    \end{subfigure}
    \hfill % Optional: add horizontal space between the subfigures
    % Second subfigure
    \begin{subfigure}{0.48\linewidth}
        \centering
        \includegraphics[width=\linewidth]{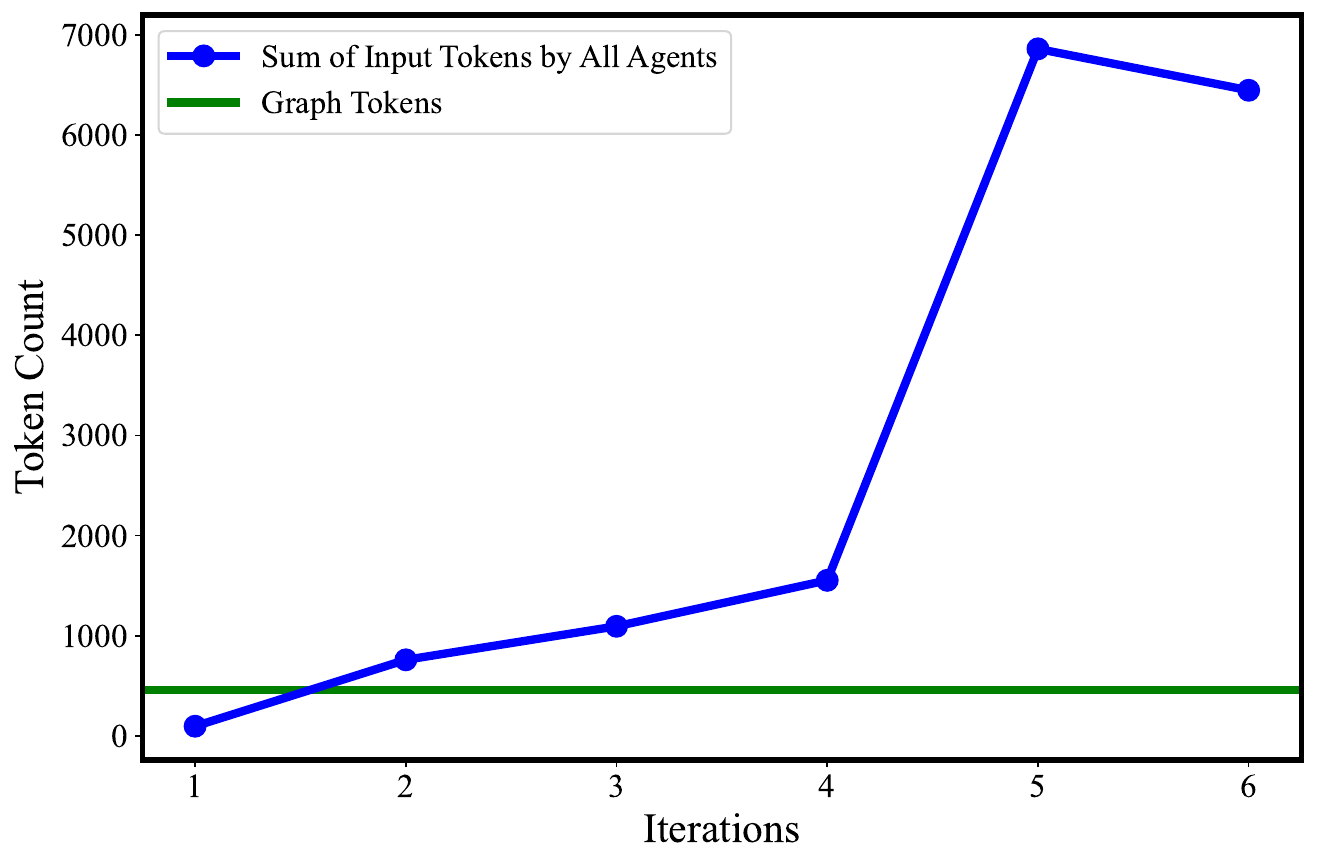}
        \caption{Trv-1}
        \label{fig:CompTrv}
    \end{subfigure}
    }
    % Overall figure caption
    \caption{\textbf{Compute Analysis}. 
    (a) NumQ\&A. For simple tasks with large graphs, \RwR takes less iterations and processes less tokens at each iteration compared to that on the graph, showing that it can efficiently filter graph information based on the task.
    (b) Trv\-1. For complex tasks with small graphs, \RwR takes more iterations and the context processed at each iteration accumulate over the graph token count, indicating that \RwR is capable of inference time scaling to address complicated tasks. 
     }
    \label{fig:CompAnlyz}
\end{figure*}

\section{Analysis on the Computational Cost}
\label{app:ComputeAnly}

We show the number of the token processed by our method (all agents) in each iteration to solve a query for the BabyAI tasks in Figure \ref{fig:CompAnlyz}. We also contrast it with the token counts of the scene graph and the CoT baseline input.
% As a direct whole-graph prompting method, the compute required by CoT is determined by the graph size. So the processed token for NumQ\&A is 4 times larger than that for the Trv-1, despite that the former is a simpler task requiring less reasoning steps.
\RwR is efficient at filtering graph information based on the task. In NumQ\&A, which is a logically simple task on a large graph,  \RwR is able to focus on key information while reasoning, resulting in less tokens processed in each iteration compared to the graph token counts.

On the other hand, \RwR is also scaling the compute for complicated tasks. In Trv-1, which is a logically more complicate task on a small graph, our method is able to establish a longer reasoning trace while dynamically shifting attention on the graph information, eventually processing all the graph data sequentially. 
This results to the more iteration and more tokens processed (including the graph token and the historical reasoning trace) compared to that in the graph.

\end{document}